# C3PA: An Open Dataset of Expert-Annotated and Regulation-Aware Privacy Policies to Enable Scalable Regulatory Compliance Audits

**Maaz Bin Musa** and **Steven M. Winston** and **Garrison Allen** and **Jacob Schiller** and **Kevin Moore** and **Sean Quick** and **Johnathan Melvin** and **Padmini Srinivasan** and **Mihailis E. Diamantis** and **Rishab Nithyanand**

University of Iowa

## Abstract

The development of tools and techniques to analyze and extract organizations' data habits from privacy policies are critical for scalable regulatory compliance audits. Unfortunately, these tools are becoming increasingly limited in their ability to identify compliance issues and fixes. After all, most were developed using regulation-agnostic datasets of annotated privacy policies obtained from a time before the introduction of landmark privacy regulations such as EU's GDPR and California's CCPA. In this paper, we describe the first open regulation-aware dataset of expert-annotated privacy policies, C3PA (CCPA Privacy Policy Provision Annotations), aimed to address this challenge. C3PA contains over *48K expert-labeled privacy policy text segments associated with responses to CCPA-specific disclosure mandates from 411 unique organizations*. We demonstrate that the C3PA dataset is uniquely suited for aiding automated audits of compliance with CCPA-related disclosure mandates.

## 1 Introduction

Privacy policies are a crucial mechanism for organizations to communicate their data habits with external entities, including consumers and regulators. Unfortunately, they have been known to fall short of meaningfully achieving these goals due to their inaccessibility or incomprehensibility (McDonald and Cranor 2009). In response, privacy advocates and researchers have focused on developing Natural Language Processing tools to make privacy policies more usable for consumers and regulators (Harkous et al. 2018; Zimmeck and Bellovin 2014).

Unfortunately, the usefulness of tools to automatically process privacy policies are hampered by the dynamicity of the digital privacy landscape. New data regulations and technologies provoke (or mandate) significant changes in the language and types of disclosures within privacy policies. These changes can limit the usefulness of tools which are developed using datasets that are unaware of these concepts.

The California Consumer Privacy Act ("CCPA," n.d.) is a hallmark privacy legislation in the United States which caused significant changes in organizations' privacy policies. Introduced in 2018 and amended in 2020, it provides comprehensive privacy protections for the data generated by Californians. The law offers Californians several rights which facilitate more control over the data that they generate. In addition, the CCPA mandates that organizations which are subject to the CCPA make 12 specific disclosures in the privacy policies. Importantly, the CCPA serves as the base framework for comprehensive privacy regulations recently introduced in other US states (Colorado, Connecticut, Delaware, Indiana, Iowa, Montana, New Jersey, Oregon, Utah, and Virginia). Consequently, the privacy regulations in these states share significant similarities with the CCPA. In fact, the 12 disclosures mandated by the CCPA are a superset of the disclosures mandated by each of them. Therefore, the ability to automatically recognize or verify the presence of CCPA-specific disclosures will serve auditors of many state regulations. Unfortunately, as we will show in this paper, existing tools are largely incapable of recognizing or verifying the presence of CCPA-mandated disclosures in privacy policies. This is because of the absence of CCPA-specific annotations during their training and fine-tuning.

This paper aims to address this gap by developing an expert-annotated dataset of CCPA-specific mandated disclosures extracted from privacy policies. To develop this dataset, we first devise a process to identify organizations that are highly likely to be subject to the CCPA and extract their privacy policies. Next, we employ a team of native-English speaking legal professionals to extract segments of text that are responsive to each of the 12 CCPA-mandated disclosures, from our dataset of privacy policies. These expert-annotations form the publicly available CCPA Privacy Policy Provision Annotations (C3PA [1]) dataset. C3PA consists of a total of 48,947 annotations extracted from 411 organizations' privacy policies. Finally, we demonstrate that tools built with C3PA data will be better suited for automatically assessing regulatory compliance with the CCPA and other similar regulations.

## 2 Related work

**Datasets.** Ramanath et al. (2014) published a dataset of 1000 privacy policies and applied Hidden Markov Models (HMM) on them to essentially create unlabeled clusters of text segments. Wilson et al. (2016) created OPP-115, a dataset of 115 privacy policies annotated using a comprehensive taxonomy of privacy concepts. The OPP-115 dataset was the first expert-annotated dataset of data practices from privacy policies and continues to be the most frequently used dataset for training models to examine privacy policies. Zimmeck et al. (2016) captured general data practices of 350 mobile app privacy policies in his dataset. Arora et al. (2022) mapped OPP-115's taxonomy to the CCPA and created a dataset of 64 privacy policies annotated with CCPA related data practices. Unfortunately, none of these datasets is specifically created using a CCPA-driven or regulation-aware taxonomy, including Arora et al. (2022). Instead, they were created to identify generic data habits and operations of organizations. Consequently, they are lacking direct connections to regulatory mandates and in fact do not contain data related to many CCPA mandates. In contrast, C3PA is the first and only fully regulation-aware expert-annotated dataset. Besides these datasets of segments from privacy policies, other work (Srinath, Wilson, and Giles 2020; Amos et al. 2021) has focused on developing tools to crawl, scrape, and parse the web to create large datasets of privacy policies. Our method for curating privacy policies borrows from these prior approaches.

[1] https://github.com/MaazBinMusa/C3PA_Dataset.git

**Privacy policy analysis.** Zimmeck and Bellovin (2014) used a Naïve Bayes classifier to create a web extension that summarizes data practices in privacy policies. Wilson et al. (2016) used their OPP-115 dataset, to train a machine learning model to predict data practices in segments of privacy policies. As a follow up, many works on privacy policy analysis leveraged OPP-115 to develop related tools. Zimmeck et al. (2019) used OPP-115 to train ML models to analyze general data practices in mobile app privacy policies of 9K apps. Sathyendra et al. (2016) and Sathyendra et al. (2017) trained ML models to detect Opt-out clauses and consent provisions in privacy policies. Since these tools are powered by regulation-unaware datasets, they are unable assess privacy policies compliance with the CCPA and other related regulations. The C3PA dataset will introduce these capabilities to existing and upcoming tools.

## 3 Annotation scheme

To analyze privacy policies that are responsive to the CCPA, it is essential to understand CCPA disclosure mandates. Table 1 lists each disclosure mandate and pairs them with the annotation labels we used. The *Updated Privacy Policy* label (L1) refers to the requirement in the CCPA section 1798.130(a)(5) that businesses display the date of the last update to their privacy policy. The *Categories* labels (L2-L4) are responsive to sections 1798.130(a)(5)(B) and (C) which require firms to disclose types of PII collected, how they are used, and with whom they are shared and sold. The *Descriptions* labels (L5-L11) are responsive to section 1798.130(a)(5)(A) which requires businesses to disclose the rights granted to consumers. Finally, the *Methods* label (L12) is responsive to section 1798.130(a)(1)(A) which requires businesses to describe how a consumer may exercise their CCPA-granted rights. It is to be noted that L6 and L10 were amendments to the CCPA made effective in 2023.

| Label | No. |
|---|---|
| Updated privacy policy | L1 |
| Categories of PI sold | L2 |
| Categories of PI shared / disclosed | L3 |
| Categories of PI collected | L4 |
| Description of right to delete | L5 |
| Description of right to correct information | L6 |
| Description of right to know PI collected | L7 |
| Description of right to know PI sold / shared | L8 |
| Description of right to opt-out of sale or sharing of PI | L9 |
| Description of right to limit use of PI | L10 |
| Description of right to non-discrimination of exercising rights | L11 |
| Methods to exercise rights | L12 |

Table 1: Annotation scheme extracted from 1798.130 section (5) of the CCPA.

## 4 Dataset creation

In this section we provide an overview of how we created the C3PA dataset. We outline our sourcing for privacy policies and our annotation process.

### 4.1 Sourcing CCPA-responsive policies

The CCPA regulations are only applicable to organizations that meet specific revenue requirements of serve a specific number of Californian users. Therefore, our approach for curating privacy policies should include related metrics. Not doing so would increase the likelihood that our analysis is focused on privacy policies which are not actually responsive to the CCPA. To address this need, we source policies from two sources: (1) Data brokers registered with the California Attorney General because they trade data obtained from a large number of Californians and (2) popular websites which are known to have trackers and a large number of Californian visitors.

**Registered data brokers**. We sourced 59% of the privacy policies in C3PA from data brokers operating within California. Data brokers are companies that specialize in collecting and selling data, often including personal information of individuals. Data brokers in California are subject to the CCPA and must register with the California Attorney General's registry[2] if they meet one of the following criteria: (1) have annual gross revenues more than twenty-five million U.S dollars, (2) handle the personal information of 50,000 or more California consumers, households, or devices or (3) derives 50 percent or more of their annual revenues from selling consumers' personal information. Data brokers in this registry self-identify as satisfying one of these criteria. We therefore expect that their privacy policies contain provisions that are responsive to the CCPA. We compiled the names and URLs of all *478 registered data brokers* from this registry as of March 2023. We refer to these organizations and associated privacy policies as *DB* in our analysis.

**Popular websites**. General popularity metrics by themselves are a poor proxy for identifying websites that are subject to the CCPA. This limitation arises due to the CCPA being pertinent only to websites that collect and share personal information of Californian residents. To address this issue, our second source consists of data from Van Nortwick and Wilson (2022) which estimated the number of unique Californian visitors and trackers on websites. From this data, we removed websites that had less than 100K unique Californian visitors per month and no trackers, leaving us with websites that are very likely to be subject to the CCPA. We selected the *top 700 websites* from this list. We refer to these organizations and associated privacy policies as *WS* in our analysis.

### 4.2 Crawler instrumentation

Next, we instrumented a crawler to locate the privacy policies associated with each organization in *DB* and *WS*.

**Identifying potential privacy policy documents**. Our crawler was developed using Playwright on Python[3]. To locate and download privacy policies, the crawler loaded the homepage of the organization and used a collection of keywords (e.g., "privacy", "notice", "CCPA", and others) to identify potential links to privacy policies. Next, the crawler clicked on each of the identified potential privacy policy links and saved their associated pages. In total, we identified 959 and 707 potential privacy policy pages for organizations in *DB* and *WS*, respectively (*cf.*, **Table 2**).

| Source | Websites | Potential policies | Similar URLs | DNSMPI URLs | 3rd party URLs | Unique content | Words (avg) |
|---|---|---|---|---|---|---|---|
| DB | 478 | 959 | 636 | 596 | 500 | 241 | 7.1K |
| WS | 700 | 707 | 700 | 665 | 393 | 170 | 10K |

**Table 2.** Summary of our post processing filtering. Each column represents the remaining privacy policies in each dataset after this step and all steps to the left were applied e.g., DNSMPI URLs column represents policies remaining after Similar URLs and DNSMPI post processing steps.

---

[2] https://oag.ca.gov/data-brokers

[3] https://playwright.dev

**Filtering unrelated and duplicate documents**. Following collection of the initial set of candidate privacy policies, we performed a series of post processing steps to ensure uniqueness and relevance of our dataset to the CCPA. (1) *Filtering out pages extracted from DNSMPI links.* The CCPA requires a "Do not share my personal information" link to be present on websites subject to it. Our heuristics-based crawler may have collected pages from such links as well. To address this, we removed any set member that had one or more keywords "do not sell", "opt out", or "opt-out" in its URL, suggesting it is a DNSMPI page. (2) *Filtering out third-party policies.* Our crawler captured all possible privacy policy, including links from third parties. Since our scope is strictly limited to the subjects of the CCPA, we removed policies obtained from third-party domains (i.e., not in *WS* or *DB*). (3) *Ensuring unique content.* Some organizations offer multiple unique URLs to the same webpage. We removed duplicate (multiple identical) policies from such organizations. Our final dataset consisted of privacy policies from *411 unique organizations* (241 in DB and 170 in WS). **Table 2** summarizes the results of each post processing step and our final dataset size.

### 4.3 Annotation procedures

Six law students who speak English as a first language annotated our 411 privacy policies. Their legal training ensures familiarity with the sort of complex, often legalistic terminology used in privacy policies. In addition to their general legal training, they were also specifically trained to understand the nuances associated with the CCPA privacy regulations.

**Annotation tooling and infrastructure.** For the annotation process we used Label-studio[4], an open-source data labeling tool. We created six amazon EC2 virtual machines, one for each annotator and ran Label-studio on each of these machines. We provided each annotator with a separate machine to avoid any interaction (and subsequent bias) between annotators. For each labeling task, the interface displayed the privacy policy text and the labels from our annotation scheme side by side as shown in Figure 2. For better readability and easier annotation experience, the privacy policies shown to each annotator had their HTML modified to: (1) remove headers, footers, and JavaScript and (2) have all tables unrolled into bullet lists.

**Annotation task.** We assigned privacy policies to annotators in batches of approximately 25 policies per week per annotator. This was to ensure that our workload never exceeded ten hours per week for each annotator. The 6[th] annotator, who assisted with logistics and team management, received only five policies per week. Each policy was *annotated by exactly three annotators.* The annotators highlighted text spans they determined were responsive to a CCPA disclosure mandate and assigned them labels from the provided list. The annotators were encouraged to highlight sentence length text spans but were not otherwise restricted. On average, an annotator took 18.5 minutes to annotate a policy. In total, our team of six expert annotators worked for 14 weeks to annotate the 411 privacy policies. To ensure all the annotators had the same understanding of their task, at the end of each week the annotators met and discussed areas of significant disagreement. In Section 4 we evaluate how this iterative process improved the quality of our dataset.

## 5 Dataset evaluation

In this section we evaluate the quality and relevance of our annotations.

### 5.1 Observed inter-annotator agreement

**Inter-annotator agreement metric.** We rely on two metrics for computing inter-annotator agreements: (1) document-level Cohen's Kappa agreement scores and (2) text-span F1 agreement scores. In the first approach, we compiled a list of annotation tags applied to each policy document by each annotator and computed the Cohen's Kappa between the two lists. This metric was found to be far too simplistic for computing meaningful agreement scores. After all, to have perfect agreement, annotators only needed to have identified the same set of disclosures within a document (i.e., which spans of text were labeled is irrelevant to this metric). To account for this gap, we also measured a text-span level F1 agreement metric as demonstrated in (Grouin et al. 2011) to preserve information granularity. **Figure 1** demonstrates how we process annotations to calculate agreement scores. In the figure, we have a document with eight words. In this example, Ann1 marks words 2-6 with the label L1, while Ann2 marks 4-8 with L1. Treating Ann1 as our reference, we have 3 true positives, 2 false positives, and 2 false negatives (yielding an F1 score of 60% for label L1 between Ann1 and Ann2). In comparison, the Kappa score would be perfect, as both annotators agree on the presence of this label in this document. The agreement between two annotators over a set of documents is the average of the F1 scores for each document. We calculated the agreement between all pairs of annotators and report the average agreement between all annotators in our dataset.

---

[4] https://labelstud.io/

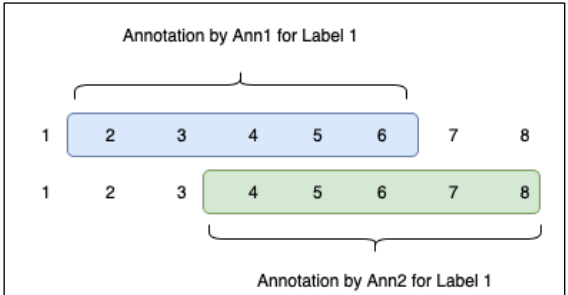

**Figure 1:** Example of how we use annotations to calculate F-1 score agreement. Here the document-level Cohen's Kappa is 1 and the text-span level F1 agreement is 60%.

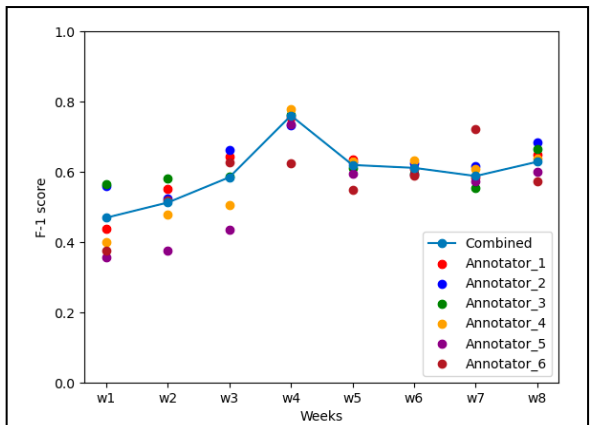

**Figure 2**: Weekly agreement scores for each annotator for the first 8 weeks. The combined agreement between all annotators gradually increases at stabilizes by week8.

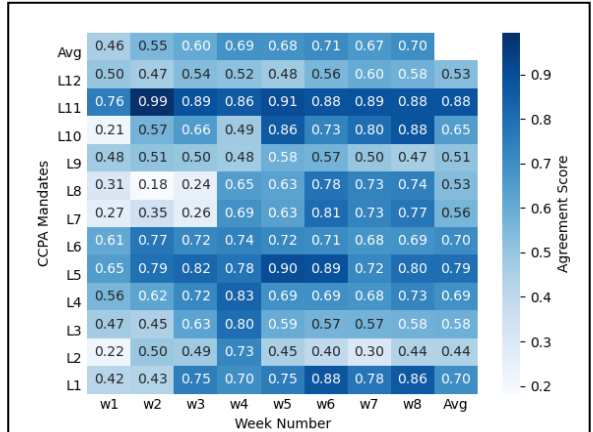

**Figure 3:** Weekly agreement scores for each mandate for the first 8 weeks. The average weekly agreement for all mandates combined starts from 0.46 in week1 and finishes at 0.70 in week8.

|  | *% of segments in APP350* | *% of segments in OPP115* |
|---|---|---|
| *L1* | 3 | 4 |
| *L2* | 0 | 0 |
| *L3* | 7 | 8 |
| *L4* | 29 | 28 |
| *L5* | 0 | 0 |
| *L6* | 0 | 0 |
| *L7* | 0 | 0 |
| *L8* | 0 | 0 |
| *L9* | 0 | 0 |
| *L10* | 0 | 0 |
| *L11* | 0 | 0 |
| *L12* | 5 | 5 |

**Table 3:** Fraction of policies from prior datasets which were classified as being responsive to a CCPA-related mandate.

**Evolution and characteristics of annotator agreement scores. Figure 2** shows agreement scores per annotator for each week during our annotation process. Agreement scores for the first three weeks increased significantly and then stabilized. Moreover, the agreement levels of each annotator with the rest of the group also stabilized towards week 8. Only the 6th annotator took more time. This is because the 6th annotator was assigned a reduced workload and hence took longer to become as adept. Figure from the Appendix shows as exactly similar trend for Kappa scores albeit the week1 and 8 scores (0.55 and 0.85) for the Kappa agreement are higher than our F1 scores. These scores suggest an increasing and stabilizing mutual understanding of the annotation task. **Figure 3** demonstrates how agreement for each label changes weekly. A clear pattern emerges based on the type of the label. The *Description* labels (L5-L11) demonstrate a high or an increasing agreement trend except for the *Description of Opt-out of sale of PI (L9)*, which has the second lowest agreement score by the end of week 8. A vigorous discussion between our annotators suggested the reason behind this trend: policies were found to describe consumer opt-out rights without explicitly mentioning any connection to the CCPA. So, the annotators had to infer a connection to the law and hence produced different annotations. Similarly, the *Categories of PI sold* and *Categories of PI shared / disclosed* (L2, L3) consistently had low agreement scores. Whereas *Categories of PI collected* (L4) had a higher agreement score because policies more often mentioned its connection to the CCPA. The *Methods* (L12) category label demonstrates a gradual but steady increase in agreement due to the annotators coming to agreement that general contact information should not be construed as a mechanism through which a consumer may make rights requests. Finally, the combined label agreement each week displays a healthy increase as the weeks progress. The score starts from a low value of 0.46 and ends at 0.70 by the end of week 8. Figure 5 from the Appendix captures an extremely high Kappa agreement score for each label per week. This is expected behavior as Kappa scores only capture agreement on the presence or absence of a label in a document, whereas our F1 score captures span-based agreements.

## 5.2 Comparison of data representativeness

We now analyze the representativeness of the C3PA dataset and available alternatives for analyzing privacy policies responsive to the CCPA. To demonstrate the unique relevance of our dataset to the CCPA, compared to previous datasets, we performed a *contextual similarity-based* analysis of annotated segments from previous datasets to CCPA mandates. Our goal was to determine how well equipped the privacy policies in previous dataset were to facilitate CCPA-related analyses.

**Analysis setup.** At a high-level, we used our annotations to develop binary classifiers associated with each CCPA mandate. We then applied these classifiers to text segments from prior datasets of privacy policies to identify passages with similar semantics as our dataset. For each CCPA mandate, we converted all of its annotations into word embeddings using the BERT sentence transformer (BERT, n.d.). We divided our embeddings into training, testing and holdout datasets (80%, 10%, 10%) and then used the training set to create 12 Logistic Regression binary classifiers which labeled a segment as being responsive to a given mandate or not (one binary classifier for each mandate). The aggregate accuracy and F-1 score for the holdout dataset was 95% and 61% respectively. Next, we gathered annotated segments from two publicly available, popular privacy policy datasets: OPP115 (5.8K segments) and APP350 (15.5K segments). Finally, we assigned labels to segments based on the outcomes of each of our 12 classifiers. In addition, we also performed a simple keyword analysis as described in Table 3 and Table 5.

**Results.** As shown by **Error! Reference source not found.** we can clearly observe the lack of contextual similarity of both previous datasets to CCPA mandates. Only 3-5% of previous datasets contain L1 (Update) and L12 (Methods) related segments even though they are somewhat generic in nature. More alarmingly, no policies from previous datasets get labeled as being responsive to L2 or L5-L11. To further bolster the uniqueness and relevance of our datasets to the CCPA, Table 4 and Table 5 in the Appendix quantify the absence of CCPA related keywords in previous datasets (and compares them to the C3PA dataset).

## 5.3 Characteristics of annotations

Our annotations process yielded a total of *48.9K annotations from 411 privacy policies*.

**Prevalence of annotations in privacy policies.** As shown in Table 4, not all labels have equal prevalence. The Categories of PI collected (L4) label appears most frequently in both DB and WS.

| Label (#segments) | Prevalence (%) | | Spread (%) | | #Words (avg.) | |
|---|---|---|---|---|---|---|
| | DB | WS | DB | WS | DB | WS |
| L1 (2.1K) | 88 | 93 | 60 | 67 | 39 | 35 |
| L2 (5.7K) | 82 | 86 | 34 | 37 | 43 | 123 |
| L3 (10.6K) | 94 | 94 | 43 | 51 | 109 | 118 |
| L4 (15.1K) | 97 | 95 | 58 | 63 | 107 | 131 |
| L5 (1.8K) | 95 | 93 | 21 | 21 | 55 | 39 |
| L6 (1.1K) | 51 | 76 | 21 | 23 | 31 | 28 |
| L7 (2.4K) | 96 | 94 | 18 | 15 | 55 | 54 |
| L8 (1.6K) | 75 | 66 | 15 | 11 | 61 | 64 |
| L9 (2.2K) | 91 | 80 | 24 | 25 | 39 | 40 |
| L10 (502) | 26 | 39 | 17 | 15 | 46 | 43 |
| L11 (1.1K) | 79 | 79 | 13 | 15 | 40 | 36 |
| L12 (4.3K) | 94 | 91 | 31 | 27 | 43 | 37 |

**Table 4.** Characteristics and distribution of CCPA mandates per policy. Specifically, the table denotes popularity and size of annotations amongst datasets. The table also captures the distance between the first and last occurrence of a mandate in a policy (spread).

In contrast, Description of the right to limit use of PI (L10) is the least frequent label in both DB and WS. These trends are consistent with our expectations as the disclosure of PI collected is a popular requirement amongst privacy regulations, whereas the right to limit use of PI is a relatively new requirement, enforced by the CPRA in January of 2023. The high prevalence of all mandates in our annotations bolsters the relevance of our dataset.

**Spread of annotations in privacy policies.** Feedback from the annotators suggested that CCPA mandates were not guaranteed to be in contiguous segments within the privacy policies. As a result, the annotators had to often skim through copious amounts of text to understand a mandate. Based on this feedback, we developed a metric to measure spread of a mandate within privacy policies. The spread % in **Table 4** captures the distance in terms of privacy policy content between the first and last occurrence of a mandate in a privacy policy. 60% spread for L1 translates to a reader needing to scan through at least 60% worth of a privacy policy text, to observe all occurrences of a mandate. Surprisingly the spread for all mandates is high, ranging from as low as 11% to as high as 67% --- suggesting the need to read a large part of a policy before understanding an organizations response to a certain mandate.

**Amount of non-CCPA related text in privacy policies.** Next, we quantified the percentage of privacy policies that are covered by non-CCPA related text. We categorize non-CCPA text as privacy policy text that was not annotated by any annotator. Our results from DB and WS suggest that 50% of their privacy policies are at most 55% and 58% covered by non-CCPA text respectively. The slight difference in values of DB and WS suggests that privacy policies from popular websites have more non-CCPA content than privacy policies from data brokers. This is expected as data brokers self-register and are more likely to be responding comprehensively and directly to the CCPA than popular websites.

## 6 Utility of the C3PA dataset

We now perform a comparative analysis between a classification model trained on OPP-115 (the most popular annotated privacy policy dataset) with a model trained on our dataset for the task of extracting CCPA-specific segments. We will show that the same model, trained on a CCPA-specific dataset (such as C3PA) will outperform one trained on a dataset which maps generic data practices to the CCPA (such as the mapping of OPP-115 to the CCPA produced by Arora et al. (2022)).

**Determining labels for the classification task.** To compare our model with a model trained on OPP-115 data, we must first create a mapping of OPP-115 labels to the CCPA mandates. As OPP's taxonomy was made before the CCPA existed, we could only map one label from it to one mandate of the CCPA (this contrasts with C3PA, which maps all CCPA disclosure mandates). The difficulty of mapping existing general taxonomies to CCPA specific mandates is consistent with previous work (cf., Section 2).

**Mapping OPP-115 labels to CCPA mandates.** We mapped the label *"first party collection/use"* from OPP's taxonomy to *"Categories of PI collected"* (*L4*) as they both capture the collection of personal information. We excluded the remaining labels for the following reasons:

- *L1 (Updated privacy policy)* was excluded as no attribute in OPP-115 category *"policy change"* captures this information.

|  | c3pa_databroker_model | | | c3pa_website_model | | | c3pa_combined_model | | |
| --- | --- | --- | --- | --- | --- | --- | --- | --- | --- |
|  | *Precision* | *Recall* | *F-1* | *Precision* | *Recall* | *F-1* | *Precision* | *Recall* | *F-1* |
| *L1* | 99 | 99 | 99 | 96 | 97 | 96 | 98 | 97 | 98 |
| *L2* | 60 | 24 | 35 | 55 | 19 | 28 | 54 | 23 | 32 |
| *L3* | 64 | 59 | 61 | 53 | 54 | 53 | 58 | 58 | 58 |
| *L4* | 70 | 86 | 77 | 62 | 85 | 72 | 68 | 83 | 75 |
| *L5* | 79 | 83 | 81 | 75 | 73 | 74 | 72 | 77 | 74 |
| *L6* | 74 | 80 | 77 | 65 | 63 | 64 | 65 | 67 | 66 |
| *L7* | 55 | 71 | 62 | 53 | 53 | 53 | 54 | 60 | 57 |
| *L8* | 61 | 35 | 45 | 57 | 30 | 39 | 61 | 30 | 40 |
| *L9* | 78 | 68 | 73 | 64 | 71 | 67 | 70 | 68 | 69 |
| *L10* | 92 | 82 | 87 | 76 | 53 | 62 | 84 | 65 | 73 |
| *L11* | 95 | 96 | 96 | 83 | 92 | 87 | 87 | 95 | 91 |
| *L12* | 77 | 85 | 81 | 78 | 76 | 77 | 82 | 85 | 83 |
| *Mac Avg* | 70 | 68 | 67 | 63 | 59 | 59 | 71 | 66 | 67 |

**Table 5:** Summary of the classification report for C3PA models. Each pair of (Precision, Recall, F-1) is marked by its model variation e.g., *c3pa_databroker_model* columns represent results from the model trained on databroker annotations and validated on website annotations. L4 is the only label comparable across c3pa and opp models.

| *Test set* | *Precision* | *Recall* | *F-1* |
| --- | --- | --- | --- |
| *website annotations* | 62 | 82 | 71 |
| *databroker annotations* | 55 | 80 | 65 |
| *combined annotations* | 60 | 80 | 68 |

**Table 6:** Classification summary for the *opp_model* on all 3 validation sets. The opp_model was trained on all annotations from the opp-115 dataset and was treated as a binary classifier for predicting L4. The table shows its performance on the three different validation sets we used for C3PA models.

- *L2 (Categories of PI sold)* and *L3 (Categories of PI shared / disclosed)* refer to sharing / disclosure / selling of PI by first-party whereas the OPP-115 categories and attributes only cover data collection / sharing practices of third parties.
- *L5-L8 (Description of right to delete / correct / know PI collected / know PI sold)* were excluded as OPP-115's category "*user access, edit and deletion*" has no attribute-value that captures sharing, selling, or deletion for attribute "*access type*" and no attribute-value of PII for attribute "*access scope*". It captures "*profile data*" and "*user account data*" instead of PII and "*view*" and "*deletion of account*" instead of sharing / selling / deletion of PII.
- *L9 (Description of right to opt-out of sale or sharing of PI)* and *L12 (Method to exercise rights)* were excluded as OPP-115's "*user choice / control*" category doesn't have an attribute-value of sharing / sold for its "*choice scope*" attribute and doesn't have any attribute-value that captures general methods for exercising rights.
- *L10 (Description of right to limit use of PI)* and *L11 (Description of right to non-discrimination)* are newer concepts to privacy and do not map to any category in OPP-115.

**Training data for a comparative evaluation.** Our first model (*opp_model*) is trained on all the annotations from OPP-115. We used the "*selected-text*" value from each annotation in OPP-115 rather than the more coarse-grained paragraph segments, as that aligns with how C3PA annotations are created. This model is trained to predict OPP-115 labels including "*first party collection/use*". For comparative analysis with the *opp_model*, we trained three variations of models using annotations from C3PA. Our first model (*c3pa_databroker_model*) uses annotations from databrokers' privacy policies for training / testing and annotations from websites' privacy policies for validation. The second model (*c3pa_website_model*) uses websites' annotations for training / testing and databrokers' annotations for validation. The third model (*c3pa_combined_model*) uses 90% of all annotations combined for training / testing and 10% of the remaining annotations for validation. For evaluating the *opp_model*, we treated it as a binary classifier where all L4 instances from the C3PA validation set(s) are labeled as positive (*"first-party collection / use"*) instances and non-L4 instances as negative instances.

**Across holdout performance.** As summarized in **Table 5** and **Table 6**, all three model variations for C3PA models show a performance increase over the *opp_model*. Interestingly, the *c3pa_databroker_model* yields the best results. This result is consistent with the description of databrokers privacy policies from section 4.1. As these policies belong to entities that self-register as being subject to the CCPA, we expect better quality annotations and as a result better performing models from these policies. This result also motivates future work assessing the tradeoffs associated with each step in the automated privacy policy analysis pipeline.

**Classification model.** To train our models, we fine-tuned the *distilbert-base-uncased* model with a *max_length=512* and *padding / truncation* enabled. We used $2e^{-5}$ *learning-rate*, *batch-size*=32 and *epochs*=2 as our training parameters. These settings are in line with the general guidelines provided for finetuning BERT models. **Table 5** and **Table 6** summarize the classification reports produced by using our validation set annotations on all models. The macro-F-1 score for *c3pa_combined_model* (67%) is similar to the final agreement scores attained by expert annotators (70%) during the annotation process. Furthermore, we also observe that *c3pa_combined_model* is more (+7%) capable of predicting L4, the common label for both models. The low scores for L2, L3 and L8 in *c3pa_combined_model* are another property that our model mimics from the annotators agreement scores and further establishes that responses to these mandates are challenging for human experts and automated tools to capture. Overall, our results highlight the effectiveness of *c3pa_combined_model* over the *opp_model*.

# 7 Concluding remarks

**Limitations.** Our analysis in this work has established the importance of utilizing regulation-aware datasets for assessing compliance (as opposed to general-purpose datasets with post-hoc mapping to regulations). However, it is unclear how representative this dataset is of privacy policies in general, or how well it will perform for general privacy policy analysis, as we focused on gathering CCPA specific policies. Finally, a systematic analysis of sources, model selection and document segmentation for building an automation tool for extracting CCPA related text from privacy policies is out of the scope of this work. However, such future work can increase the privacy policies that can be analyzed efficiently and further our understanding of CCPA rights and requirements.

**Conclusion.** In this work we developed a first of its kind dataset of privacy policies annotated with CCPA mandates. Our dataset comprises 411 privacy policies annotated by six domain experts using an annotation scheme derived from the CCPA. Our analysis demonstrates how our dataset stands out from previous datasets in terms of relevance to modern privacy policies drafted in the shadow of the CCPA. We go one step further and showcase how our unique dataset surpasses previous datasets in powering tools for CCPA analysis. While previous datasets were crucial in understanding general privacy practices in privacy policies prior to 2018, our dataset improves the capabilities of models to assess compliance with regulations based on, or sharing similarities with, the CCPA. The C3PA dataset is available at: https://github.com/MaazBinMusa/C3PA_Dataset.git.

# 8 Acknowledgements

This material is based upon work supported by the National Science Foundation under Grant Nos. CNS-2335659 and CNS-2338377. Any opinions, findings, and conclusions or recommendations expressed in this material are those of the authors and do not necessarily reflect the views of the National Science Foundation.

# 9 Appendix

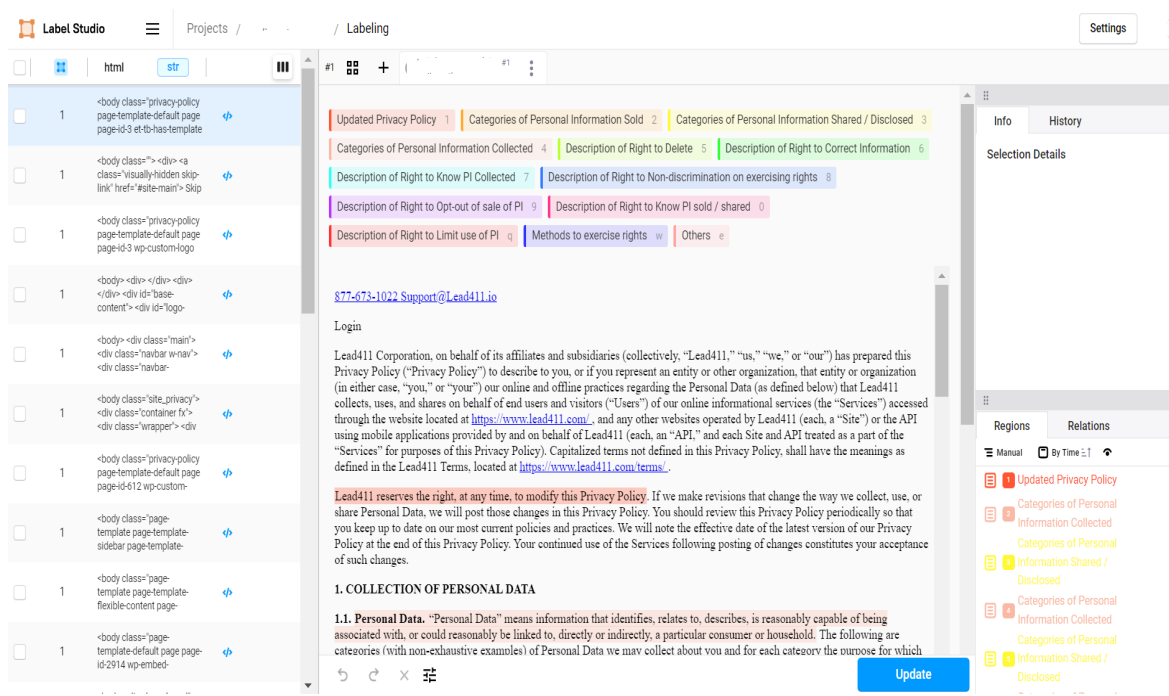

Figure 2: Layout of our label studio annotation instance. The cleaned-up privacy policy html is displayed in the center with the list of labels on the top.

| Right | Keywords |
|---|---|
| General | ccpa, california consumer, california privacy |
| Right to Delete | delete, deletion, deleted, deleting |
| Right to Know | know, knowing |
| Right to non-discrimination | non discrimination, non-discrimination, discrimination, discriminate, discriminating, discriminatory |

Table 4: We create a regular expression that finds phrases that start with the word "right" and end with any of the keywords from the table. The regular expression also allows 0-4 words between the word "right" and any of the keywords.

| Dataset | Total Privacy Policies | CCPA Relevance | Right to Delete | Right to Know | Right to Non-discrimination | All |
|---|---|---|---|---|---|---|
| **APP350** | 349 | 37 (11%) | 9 (3%) | 13 (4%) | 0 (0%) | 0 (0%) |
| **OPP115** | 115 | 41 (36%) | 2 (2%) | 2 (2%) | 0 (0%) | 0 (0%) |
| **MAPP350** | 64 | 39 (61%) | 24 (38%) | 9 (14%) | 9 (14%) | 0 (0%) |
| **DB** | 241 | 241 (100%) | 196 (81%) | 144 (60%) | 114 (47%) | 9 (4%) |
| **WS** | 170 | 170 (100%) | 112 (72%) | 94 (60%) | 80 (51%) | 9 (6%) |

Table 5: Keyword analysis on the privacy policies in various datasets. The 'CCPA relevance' column captures how many privacy policies mention the CCPA. The 'Right to' columns capture mention of specific CCPA rights, whereas the 'All' column captures the presence of all rights given to users.

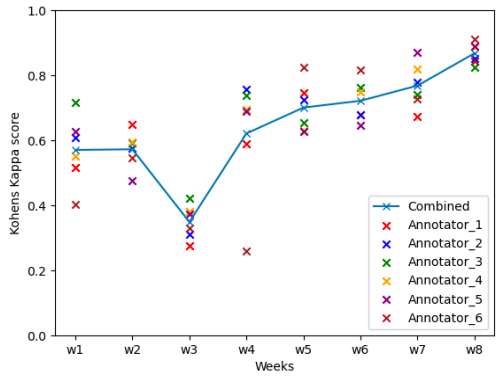

Figure 4: Weekly agreement (Kappa) scores for each annotator for the first 8 weeks.

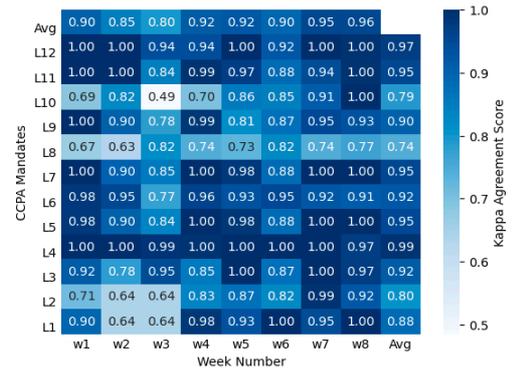

Figure 5: Weekly agreement (Kappa) scores for each mandate for the first 8 weeks. The average weekly agreement for all mandates combined starts from 0.90 in week1 and finishes at 0.96 in week8.

13